\documentclass[lettersize,journal]{IEEEtran}
\usepackage{amsmath,amsfonts}
\usepackage{algorithmic}
\usepackage{algorithm}
\usepackage{array}
\usepackage[caption=false,font=normalsize,labelfont=sf,textfont=sf]{subfig}
\usepackage{textcomp}
\usepackage{stfloats}
\usepackage{url}
\usepackage{verbatim}
\usepackage{graphicx}
\usepackage{cite}

\usepackage{bm,booktabs,mathtools,cancel,hyperref,multirow}
\usepackage[table,dvipsnames]{xcolor} 
\usepackage[normalem]{ulem}

\begin{document}

\title{A Survey of Behavior Foundation Model: Next-Generation \\Whole-Body Control System of Humanoid Robots}


\author{Mingqi Yuan\textsuperscript{123*}, Tao Yu\textsuperscript{2*}, Wenqi Ge\textsuperscript{24*}, Xiuyong Yao\textsuperscript{2}, Dapeng Li\textsuperscript{2}, Huijiang Wang\textsuperscript{5}, Jiayu Chen\textsuperscript{48}, Bo Li\textsuperscript{1}, Wei Zhang\textsuperscript{26},~\IEEEmembership{Senior Member,~IEEE}, Wenjun Zeng\textsuperscript{3},~\IEEEmembership{Fellow,~IEEE}, Hua Chen\textsuperscript{27$\ddagger$}, Xin Jin\textsuperscript{3$\dagger$}
\thanks{
\textsuperscript{1}Department of Computing, The Hong Kong Polytechnic University, Hong Kong SAR, China. \textsuperscript{2}LimX Dynamics, Shenzhen, China. \textsuperscript{3}Ningbo Institute of Digital Twin, Eastern Institute of Technology, Ningbo, China. \textsuperscript{4}Department of Data and Systems Engineering, The University of Hong Kong, Hong Kong SAR, China. \textsuperscript{5}CREATE Lab, EPFL, Lausanne, Switzerland. \textsuperscript{6}School of System Design and Intelligent Manufacturing, Southern University of Science and Technology, Shenzhen, China. \textsuperscript{7}ZJU-UIUC Institute, Zhejiang University, Zhejiang, China. \textsuperscript{8}INFIFORCE Intelligent Technology Co., Ltd., Zhejiang, China.
}
\thanks{\textsuperscript{*}Work done at LimX Dynamics, and these authors contributed equally.}
\thanks{\textsuperscript{$\dagger$}Corresponding author: Xin Jin {\tt\small(jinxin@eitech.edu.cn)}.}
\thanks{\textsuperscript{$\ddagger$}Project lead: Hua Chen {\tt\small(huachen@intl.zju.edu.cn)}.}
}

\markboth{IEEE Transactions on Pattern Analysis and Machine Intelligence}%
{Shell \MakeLowercase{\textit{et al.}}: A Sample Article Using IEEEtran.cls for IEEE Journals}


\maketitle

\begin{abstract}
Humanoid robots are drawing significant attention as versatile platforms for complex motor control, human-robot interaction, and general-purpose physical intelligence. However, achieving efficient whole-body control (WBC) in humanoids remains a fundamental challenge due to sophisticated dynamics, underactuation, and diverse task requirements. While learning-based controllers have shown promise for complex tasks, their reliance on labor-intensive and costly retraining for new scenarios limits real-world applicability. To address these limitations, behavior(al) foundation models (BFMs) have emerged as a new paradigm that leverages large-scale pre-training to learn reusable primitive skills and broad behavioral priors, enabling zero-shot or rapid adaptation to a wide range of downstream tasks. In this paper, we present a comprehensive overview of BFMs for humanoid WBC, tracing their development across diverse pre-training pipelines. Furthermore, we discuss real-world applications, current limitations, urgent challenges, and future opportunities, positioning BFMs as a key approach toward scalable and general-purpose humanoid intelligence. Finally, we provide a curated and regularly updated collection of BFM papers and projects to facilitate further research, which is available at \url{https://github.com/yuanmingqi/awesome-bfm-papers}.
\end{abstract}


\section{Introduction}
\IEEEPARstart{H}{umanoid} robots are increasingly being developed and deployed in a variety of real-world scenarios due to their human-like morphology and high degrees of freedom (DoF) \cite{darvish2023teleoperation,valenzuela2024embodying,tong2024advancements}. These capabilities allow them to operate seamlessly in environments originally designed for humans, allowing them to perform locomotion, manipulation, and interaction tasks with versatility and agility. To enable such rich behaviors, whole-body control (WBC) plays a central role by coordinating numerous joints, contacts, and actuators to achieve balanced, agile, and safe full-body motions. However, designing effective WBC policies remains highly challenging due to the complex dynamics of humanoids, underactuation, frequent contact changes, and the strong coupling between locomotion, manipulation, and balance \cite{kuindersma2016optimization,cheng2024expressive}. These difficulties highlight the need for more general, robust, and scalable control frameworks. In the following sections, we begin by comprehensively reviewing previous humanoid WBC methods, ranging from traditional model-based to learning-based and task-specific controllers, before introducing the transformative approach---the behavior foundation model. This evolutionary trajectory not only reflects the field's advancement toward enhanced intelligence and generalizability but also paves the way for the next-generation humanoid robot control systems.

\begin{figure}[h!]
    \centering
    \includegraphics[width=\linewidth]{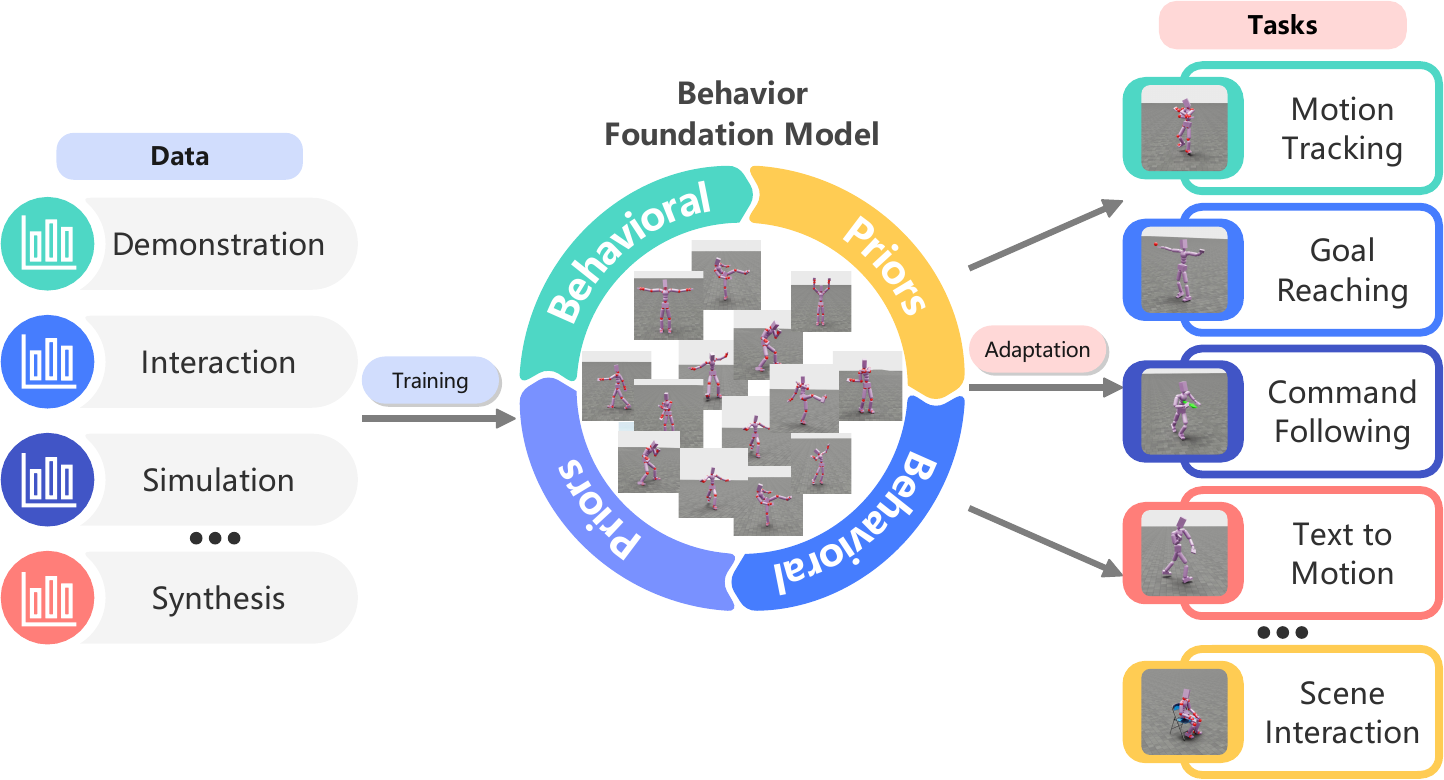}
    \caption{A behavior foundation model learns broad behavior priors from large-scale and diverse behavior data, which can then be conveniently adapted to a wide range of downstream tasks.}
    \label{fig:bfm preface}
\end{figure}

\subsection{Traditional Model-based Controller}

Traditional WBC methods have served as the cornerstone for locomotion and manipulation in early humanoid robots \cite{kulic2016anthropomorphic,goswami2019humanoid}. These methods rely heavily on physics-based models and are typically structured into a predictive-reactive hierarchy: high-level planners like centroidal model predictive control (MPC) \cite{schwenzer2021review,romualdi2022online} generate reference trajectories, while low-level task-space whole-body controllers solve optimal control problems (OCPs) \cite{sethi2021optimal} to track these objectives under dynamic constraints. For instance, operational space control \cite{khatib2003unified} and hierarchical task control \cite{sentis2005synthesis} establish the theoretical foundation, while \cite{escande2014hierarchical} advances hierarchical quadratic programming (QP) solvers and enables real-time performance in multi-task scenarios such as balance, walking, and manipulation. Whole-body operational space control (WBOSC) extends these methods by integrating the control of multiple body parts, handling redundancy, and managing dynamic constraints across the entire robot body \cite{sentis2006whole,kim2016stabilizing,fok2016controlit}. These frameworks have been widely applied in humanoids, such as Atlas \cite{kuindersma2016optimization}, HRP-2 \cite{10000129}, and DLR's torque-controlled robots \cite{henze2015approach}, achieving robust locomotion and multi-contact interactions. 

Despite their success, traditional WBC systems face critical limitations: (i) task design, gain tuning, and heuristic adjustments for complex behaviors (\textit{e.g.}, uneven terrain or dynamic transitions) remain labor-intensive and brittle, (ii) real-time MPC struggles with high-dimensional systems, often requiring simplifications that sacrifice dynamic fidelity  \cite{sovukluk2023whole}, (iii) the lack of flexibility to execute highly dynamic skills (\textit{e.g.}, backflips or rapid contact switches) or adapt to unforeseen disturbances \cite{ishihara2019full}, and (iv) weak robustness, as even a little push may topple a robot with model-based walking controller. These challenges are especially significant in humanoids, where tasks often require rich coordination, contact reasoning, and situational awareness \cite{hao2025embodied, murooka2024whole}. As a result, recent research increasingly shifts toward data-driven approaches, aiming to learn motor skills, coordination policies, and behavioral priors from demonstrations or reinforcement learning \cite{peng2018deepmimic,singh2021parrot}. 

\begin{figure*}[ht!]
    \centering
    \includegraphics[width=\linewidth]{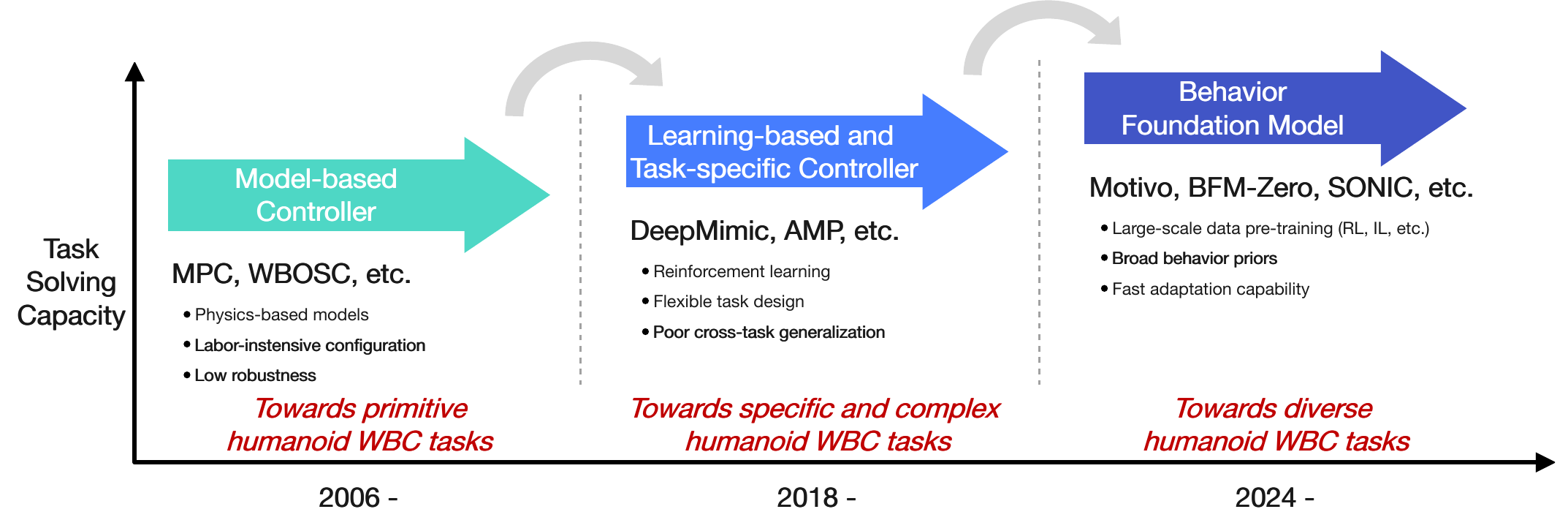}
    \caption{Evolution map of the whole-body controller for humanoid robots.}
    \label{fig:emap}
\end{figure*}

\subsection{Learning-based and Task-specific Controller}

Learning-based methods, particularly reinforcement learning (RL) and imitation learning (IL), have emerged as promising alternatives to traditional WBC methods, enabling robots to acquire complex skills through environmental interaction or human demonstrations \cite{kober2013reinforcement,sunderhauf2018limits,nguyen2019review,karoly2020deep,liu2021deep,huang2022reward,zhang2022adjacency,wu2023human,chen2023hierarchical,zhu2023transfer,luo2024moil}. For example, the DeepMimic framework \cite{peng2018deepmimic} combines deep RL with motion capture data, allowing physically simulated characters to learn dynamic skills while maintaining natural motion quality. This work is further extended by adversarial motion priors (AMP) \cite{peng2021amp}, which enables more stylized and diverse character control while preserving physical realism. Similarly, the HoST \cite{huang2025learning} framework introduces an RL-based standing-up controller that learns adaptive and stable motions across diverse laboratory and outdoor environments, highlighting the robustness and task-specific learning capability of RL. For IL-based approaches, the TRILL \cite{seo2023deep} approach combines virtual reality (VR) teleoperation with WBC for humanoid loco-manipulation, demonstrating an 85\% success rate in real-world bimanual tasks \cite{bang2023control}. Further advances, such as ExBody \cite{cheng2024expressive,ji2025exbody2}, develop an expressive WBC framework that decouples upper-body IL (for stylistic motions) from robust lower-body locomotion, enabling humanoid robots to dynamically adapt their gait while performing diverse movements. This approach overcomes the instability of full-body imitation caused by morphological mismatches between humans and robots.

While learning-based methods have demonstrated remarkable success in diverse humanoid WBC tasks, they face fundamental challenges that limit their broader applicability. RL-based approaches suffer from sample inefficiency, often requiring millions of environment interactions to converge, while remaining highly sensitive to the reward function design---poorly shaped rewards can lead to unintended behaviors or local optima \cite{maeureka}. Furthermore, the simulation-to-reality (Sim2Real) gap exacerbates these limitations, as policies trained in simulation frequently degrade when confronted with real-world dynamics, sensor noise, and hardware imperfections \cite{kadian2020sim2real,hofer2021sim2real,iyengar2023sim2real,su2024sim2real}. In contrast, IL-based methods are more sample-efficient yet pose a significant challenge to data collection, and learned policies often inherit the biases and limitations of the demonstrator \cite{schaal1999imitation,hussein2017imitation,laskey2017dart,fang2019survey,hua2021learning}. Moreover, both paradigms struggle with generalization, in which learned policies typically excel only at narrow tasks and fail to adapt to new scenarios without extensive retraining. These challenges collectively underscore the need for approaches that combine the flexibility of learning with structured priors for robustness and generalizability, a gap that behavior foundation models aim to bridge.

\subsection{Behavior Foundation Model}\label{sec:intro-bfm}
The term "\textbf{behavior(al) foundation model (BFM)}" is first introduced in \cite{pirotta2024fast}, which proposes a successor measure-based framework for training generalist policies capable of instantly imitating diverse behaviors from minimal demonstrations. It demonstrates that BFMs pretrained on unsupervised interaction data are promising for eliminating task-specific RL fine-tuning by solving imitation tasks through forward-backward state feature matching, while simultaneously supporting multiple IL paradigms via a unified representation, such as behavioral cloning, reward inference, and distribution matching. Subsequent work \cite{cetin2025finer,tirinzoni2025zeroshot,sikchi2025fast,vainshtein2025task,bobrin2025zero} has established BFMs as a class of RL agents capable of unsupervised training on reward-free transitions while yielding approximately optimal policies for broad classes of reward functions at test time without additional learning or planning.

In this paper, we extend the definition of BFMs as a specialized class of foundation models \cite{bommasani2021opportunities} designed to control agent behaviors in dynamic environments. As illustrated in Fig.~\ref{fig:bfm preface}, rooted in the principles of general foundation models (\textit{e.g.}, GPT-4 \cite{achiam2023gpt}, CLIP \cite{radford2021learning}, and SAM \cite{kirillov2023segment}) that leverage broad, self-supervised pre-training on large-scale static data, BFMs are often trained on extensive behavior data (\textit{e.g.}, human demonstrations, or agent-environment interactions), encoding a comprehensive spectrum of behaviors rather than specializing narrowly in single-task scenarios. This property ensures that the model can readily generalize across different tasks, contexts, or environments, demonstrating versatile and adaptive behavior generation capabilities. Recent advancements in vision-language-action (VLA) models \cite{radford2021learning,zitkovich2023rt,kim2024openvla,wen2025tinyvla} have focused on integrating vision, language, and action to handle multi-modal tasks, excelling in dynamic settings where they generate context-aware responses based on visual and linguistic inputs. In contrast, BFMs are primarily designed for directly controlling agent behaviors such as locomotion, manipulation, and interaction. Moreover, most existing VLA models are applicable to relatively stable platforms, such as mechanical arms or wheeled humanoid robots \cite{zambella2019dynamic}, whereas BFMs are designed to handle the sophisticated WBC of full-size humanoid robots.

Inspired by the discussions above, it is worthwhile to conduct a systematic and comprehensive review of BFMs to provide a holistic perspective for subsequent research. To the best of our knowledge, this is the first study focused on the development of BFMs, particularly their applications in humanoid robots. The structure of this paper is organized as follows: Section~\ref{sec:bg} introduces the essential background information of this paper. Section~\ref{sec:bfm robot} discusses the application of BFMs in humanoid WBC, including diverse pre-training and adaptation strategies. Section~\ref{sec:bfm app limit} explores the potential applications of BFMs towards multiple industries, while summarizing the limitations of current BFMs. Section~\ref{sec:bfm oppo risks} highlights the opportunities for future advancements in BFMs, as well as the risks and ethical concerns associated with their development and deployment. Finally, Section~\ref{sec:conclusion} summarizes the key findings and contributions of this paper.

\section{Background}\label{sec:bg}
In this section, we introduce the basic background to support the subsequent analysis of BFM approaches. More specifically, we begin with an overview of the definition and evolution of the humanoid WBC systems. Then, we introduce the formulation of RL, which is currently widely employed to build learning-based and task-specific controllers. 

\subsection{Humanoid Whole-body Control}
Humanoid robots are expected to operate in unstructured environments that are dynamic and unpredictable, demanding control systems that are highly versatile, robust, reconfigurable, dexterous, and mobile compared to less agile robotic platforms \cite{moro2019whole}. To that end, the humanoid WBC is proposed to coordinate the motion of multiple robots' appendages to execute multiple tasks simultaneously and reliably. It considers the entire robot body as a single and integrated system, managing locomotion, manipulation, and interaction with the environment using a unified set of control algorithms \cite{sentis2006whole}. As illustrated in Fig.~\ref{fig:emap}, the humanoid WBC has evolved from traditional model-based approaches to flexible learning-based approaches, moving toward a generalist that solves broad tasks in diverse scenarios \cite{moro2019whole,zambella2019dynamic,cheng2024expressive,xue2025unified,he2025hover,wang2025experts}. In line with this trend, BFMs have emerged as a promising approach to achieve general-purpose WBC through large-scale pre-training on diverse motion data, which will be comprehensively discussed in the following contents.

\subsection{Reinforcement Learning}
We frame the RL problem as a Markov decision process (MDP), which can be defined by a tuple $\mathcal{M} = (\mathcal{S}, \mathcal{A}, P, r^{\rm ext}, \gamma)$ \cite{sutton1998reinforcement}. Here, $\mathcal{S}$ is the state space, $\mathcal{A}$ is the action space, $P(\bm{s}'|\bm{s},\bm{a})$ is the probability measure on $\mathcal{S}$ defining the stochastic transition to the next state $\bm{s}'$ obtained by taking action $\bm{a}$ in state $\bm{s}$, $r^{\rm ext}(\bm{s}):\mathcal{S}\rightarrow\mathbb{R}$ is the extrinsic/task reward function, and $\gamma \in [0,1]$ is a discount factor. A policy $\pi$ is defined as the probability measure $\pi(\bm{a}|\bm{s})$ that maps each state to a distribution over actions. Furthermore, we denote $\mathrm{Pr}(\cdot|\bm{s}_0,\bm{a}_0,\pi)$ and $\mathbb{E}(\cdot|\bm{s}_0,\bm{a}_0,\pi)$ as the probability and expectation operators under state-action sequences $(\bm{s}_t,\bm{a}_t)_{t\geq 0}$ starting at $(\bm{s}_0,\bm{a}_0)$ and following the policy $\pi$ with $\bm{s}_t\sim P(\bm{s}_t|\bm{s}_{t-1},\bm{a}_{t-1})$ and $\bm{a}_t \sim \pi(\bm{a}_t|\bm{s}_t)$. The goal of RL is to learn a policy that maximizes the expected discounted return:
\begin{equation}\label{eq:rl}
J_{\pi}=\mathbb{E}_{\pi}\left[\sum_{t=0}^{\infty}\gamma^{t}r^{\rm ext}(\bm{s}_{t})\right].
\end{equation}

\begin{table}[t!]
\centering
\caption{Main notations and their semantics used in this article.}
\label{tb:notation}
\begin{tabular}{c|l}
\toprule
\textbf{Notation} & \textbf{Semantics} \\ \midrule
    $\mathcal{M}$     &   Markov decision process       \\
    $\mathcal{S}$     &   State space        \\
    $\mathcal{A}$     &   Action space        \\
    $r^{\rm ext}(\bm{s})$ & Extrinsic/task reward function \\
    $r^{\rm int}(\bm{s})$ & Intrinsic reward function \\
    $P(\mathrm{d}\bm{s}'|\bm{s},\bm{a})$     & Transition probability measure          \\
    $\gamma$ & Discount factor \\
    $\pi(\mathrm{d}\bm{a}|\bm{s})$ & Policy \\
    $Q^{\pi}_{r}(\bm{s},\bm{a})$     &  Action-value function of policy $\pi$ and reward $r$         \\
    $M^{\pi}(\mathcal{X}| \bm{s}, \bm{a})$      &   Successor measure of policy $\pi$       \\
    $F(\bm{s}, \bm{a})$     & Forward embedding          \\
    $B(\bm{s}')$     & Backward embedding          \\
    $L$ & Loss function \\\bottomrule
\end{tabular}
\end{table}

Finally, we list all the main notations and their semantics used in this paper in Table~\ref{tb:notation}.

\begin{figure*}[ht!]
\centering
\includegraphics[width=\linewidth]{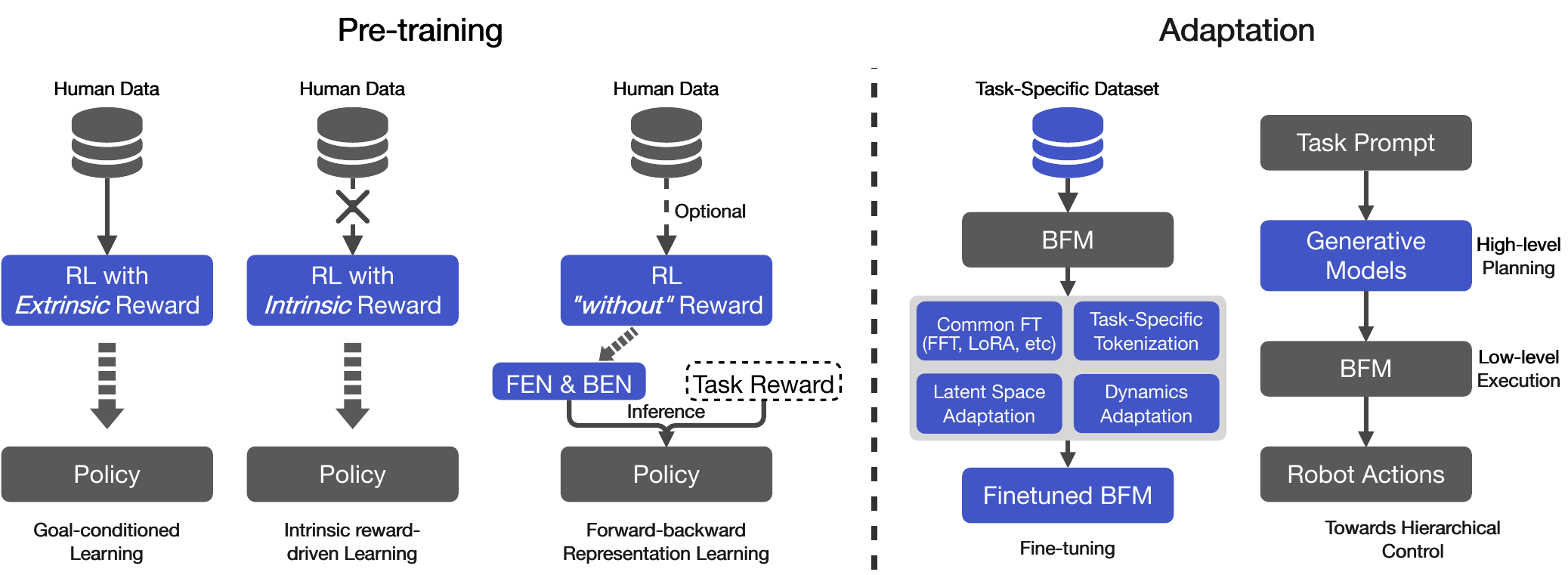}
\caption{An overview of the pre-training pipelines and adaptation strategies for BFMs discussed in this review. The goal-conditioned learning requires an \textbf{extrinsic reward} function and large-scale human data, while intrinsic reward-driven learning uses \textbf{intrinsic rewards} generated by self-supervised tasks. In contrast, the forward-backward representation learning learns a forward embedding network (FEN) and a backward embedding network (BEN) using \textbf{reward-free} transitions, which can then be combined with a specific reward function to infer a policy. For adaptation strategies, BFMs can be fine-tuned through common approaches, such as full fine-tuning (FFT) and low-rank adaptation (LoRA), as well as methods like latent space adaptation, which adjust the policy by modifying the latent task vector. Beyond fine-tuning, adaptation also involves strategies for hierarchical control, where high-level planners (\textit{e.g.}, generative models like LLMs or diffusion models) process abstract goals and generate subtasks for the BFM to execute as a low-level controller, enabling complex and long-horizon task completion.}
\label{fig:taxonomy}
\end{figure*}

\section{BFM for Humanoid Whole-body Control}\label{sec:bfm robot}

In this section, we introduce representative methods for constructing and adapting BFMs towards humanoid WBC, analyzing their main motivations, typical implementations, and empirical properties extensively. Fig.~\ref{fig:taxonomy} offers an overview of the pre-training pipelines and adaptation strategies that will be discussed in this review, and Table~\ref{tb:taxonomy} lists the corresponding methods. \textbf{Note that some approaches might have been proposed before the emergence of the concept of BFMs, yet we still discuss them as long as they adhere to the properties of BFMs or their physical meaning is analogous to BFMs.}

\subsection{Pre-training}
Pre-training of BFMs seeks to learn reusable primitive skills and behavioral priors from large-scale data sources, creating a foundation for efficient downstream adaptation. Current approaches can be broadly categorized into three types: \textbf{goal-conditioned learning}, \textbf{intrinsic reward-driven learning}, and the \textbf{forward-backward representation learning}.

\subsubsection{Goal-conditioned Learning}

\begin{figure}[h!]
    \centering
    \includegraphics[width=1.\linewidth]{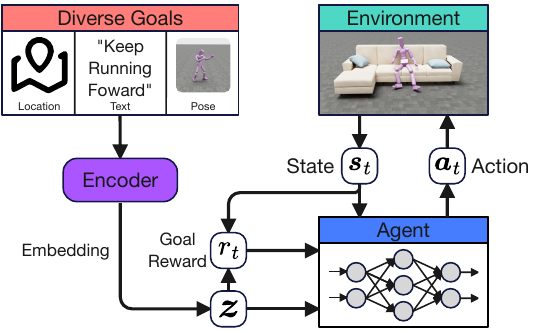}
    \caption{Workflow of the goal-conditioned learning, which enables versatile skill acquisition by training policies to achieve diverse target states specified through goal embeddings.}
    \label{fig:goal}
\end{figure}

As shown in Figure~\ref{fig:goal}, goal-conditioned learning is a framework in RL where an agent's behavior is conditioned on a specific goal or objective, typically provided as input. Unlike traditional RL, where the agent learns from raw state-action pairs without explicit task-specific guidance, goal-conditioned learning integrates the goal into the agent's policy, enabling it to adapt its actions toward achieving that specific goal. The goal can be specified in various forms, such as a target state, an objective function, or an external task description \cite{luo2023perpetual,wu2023masked,jiang2024hgap}. The key advantage of goal-conditioned learning lies in its ability to learn a more flexible and transferable policy that can be applied to a wide range of tasks, as it directly incorporates the task's goal during training, rather than requiring retraining for each specific task. This makes it particularly useful in environments where the agent needs to solve multiple tasks or interact with changing environments.

\begin{table*}[ht!]
\centering
\normalsize
\caption{Taxonomy of the pre-training and adaptation approaches of the BFMs. The "*" symbol indicates that this work has been successfully deployed in real-world humanoid robots.}
\label{tb:taxonomy}
\begin{tabular}{ccc}
\toprule[1.0pt]
\rowcolor{gray!20} \textbf{Item}                                                                                                             & \textbf{Category}                                                                  & \textbf{Algorithm}                                                                                                                                                                                                                                                                                                                                                                                                                                \\ \midrule[1.0pt]
\textbf{}                                                                                                                                    & \begin{tabular}[c]{@{}c@{}}Goal-Conditioned\\ Learning\end{tabular}                & \begin{tabular}[c]{@{}c@{}}TeamPlay \cite{liu2022motor}, ASE \cite{peng2022ase}, PHC \cite{luo2023perpetual}, CALM \cite{tessler2023calm}, MTM \cite{wu2023masked},\\ CASE \cite{dou2023case}, InterMimic \cite{xu2025intermimic}, MoConVQ \cite{yao2024moconvq}, H-GAP \cite{jiang2024hgap},\\ MaskedMimic \cite{tessler2024maskedmimic}, HugWBC$^*$ \cite{xue2025unified}, HOVER$^*$ \cite{he2025hover}, ModSkill \cite{huang2025modskill},\\ CLONE$^*$ \cite{li2025clone}, TWIST$^*$ \cite{ze2025twist}, TWIST2$^*$ \cite{ze2025twist2}, AMS$^*$ \cite{pan2025agility},\\ Any2Track$^*$ {\cite{zhang2025track}, BFM4Humanoid$^*$ \cite{zeng2025behavior}, SONIC$^*$ \cite{luo2025sonic}}\end{tabular} \\ \cmidrule{2-3} 
\textbf{Pre-training}                                                                                                                        & \begin{tabular}[c]{@{}c@{}}Intrinsic Reward-Driven\\ Learning\end{tabular}         & \begin{tabular}[c]{@{}c@{}}ICM \cite{pathak2017curiosity}, DIAYN \cite{eysenbach2018diversity}, RND \cite{burda2018exploration}, APS \cite{liu2021aps},\\ ProtoRL \cite{yarats2021reinforcement}, RE3 \cite{seo2021state}, ODPP \cite{chen2023unified}\end{tabular}                                                                                                                                                                               \\ \cmidrule{2-3} 
                                                                                                                                             & \begin{tabular}[c]{@{}c@{}}Forward-Backward\\ Representation Learning\end{tabular} & \begin{tabular}[c]{@{}c@{}}FB \cite{touati2021learning}, FB-IL \cite{pirotta2024fast}, FB-AW \cite{cetin2025finer}, FB-ARE \cite{cetin2025finer}, \\FB-AWARE \cite{cetin2025finer}, Motivo \cite{tirinzoni2025zeroshot}, BFM-Zero$^*$ \cite{li2025bfm}\end{tabular}                                                                                                                                                                                                                                                                                                                    \\ \midrule
\multirow{2}{*}{\textbf{\begin{tabular}[c]{@{}c@{}}\textcolor{white}{Adaptation}\\ Adaptation\end{tabular}}} & \begin{tabular}[c]{@{}c@{}}Fine-tuning\\ Techniques\end{tabular}                   & \begin{tabular}[c]{@{}c@{}}TokenHSI \cite{pan2025tokenhsi}, Belief-FB \cite{bobrin2025zero}, Rotation-FB \cite{bobrin2025zero},\\ TaskTokens \cite{vainshtein2025task}, ReLA \cite{sikchi2025fast}, LoLA \cite{sikchi2025fast}\end{tabular}                                                                                                                                                                                                       \\ \cmidrule{2-3} 
                                                                                                                                             & \begin{tabular}[c]{@{}c@{}}Towards Hierarchical\\ Control\end{tabular}             & \begin{tabular}[c]{@{}c@{}}UniHSI \cite{xiao2024unified}, TokenHSI \cite{pan2025tokenhsi}, UniPhys \cite{wu2025uniphys}, CLoSD \cite{tevet2025closing},\\LangWBC$^*$ \cite{shao2025langwbc}, LeVERB$^*$ \cite{xue2025leverb}, BeyondMimic$^*$ \cite{liao2025beyondmimic}, SENTINEL$^*$ \cite{wang2025sentinel}\end{tabular}                                                                                                                                                                                                                \\ \bottomrule[1.0pt]
\end{tabular}
\end{table*}

\noindent\textbf{Skill learning from motion tracking}. Among the diverse approaches to goal-conditioned learning, tracking-based learning represents a specialized form where the target behavior is explicitly defined by dense reference supervision or guidance, typically derived from motion capture data or expert demonstrations. At each time step, the agent is often trained to track the given reference motion's joint angle or kinematic pose of the next time step \cite{peng2018deepmimic}. The primary motivation behind tracking-based learning is that learning to track a single pose is more achievable and general than directly imitating a whole motion, especially a complex motion.

For example, the TeamPlay \cite{liu2022motor} framework trains an agent to imitate a large amount of football motion capture data via a DeepMimic-like \cite{peng2018deepmimic} approach, aiming to learn a complete behavior set required for the football game. Then the agent is leveraged to sample substantial state-action pairs to train a neural probabilistic motor primitive (NPMP) model \cite{merel2018neural} and derive a low-level latent-conditioned controller. Finally, the controller is applied for further drill learning by conducting RL with a drill-specific (\textit{e.g.}, follow, dribble, shoot, and kick-to-target) reward function. Here, the learned low-level controller can be viewed as a BFM as it learns realistic human-like movements based on motion capture data and can be rapidly adapted to diverse higher-level drill learning. 

Similarly, the adversarial skill embeddings (ASE) framework \cite{peng2022ase} learns a reusable latent space of motor skills by combining adversarial IL with unsupervised RL. Trained on unstructured motion data, ASE produces a latent-conditioned low-level controller capable of generating diverse and physically plausible behaviors, serving as a general-purpose motor prior to downstream tasks. Building on ASE, the conditional adversarial latent models (CALM) framework \cite{tessler2023calm} incorporates a conditional discriminator to enable fine-grained control over generated motions through latent manipulation. This line of work is further extended by CASE \cite{dou2023case}, which introduces skill-conditioned IL with training techniques such as focal skill sampling and skeletal residual forces to enhance agility and motion diversity.

While the methods above achieve efficient skill acquisition from large behavior datasets, the HugWBC \cite{xue2025unified} method explores learning versatile locomotion skills without relying on pre-collected motion data. The framework automatically generates adaptive behaviors through a structured RL process, where a general command space dynamically produces feasible velocity, gait, and posture targets during training. By reformulating WBC as a self-supervised command-tracking problem, the work establishes a new direction for developing general-purpose humanoid controllers that learn robust skills through environmental interaction rather than data imitation.

Moving beyond body-level skill learning, ModSkill \cite{huang2025modskill} introduces a modular framework that decouples full-body motion into part-specific skills for individual body parts. This modularization allows for efficient and scalable learning, as each body part is controlled independently by a low-level controller driven by part-specific skill embeddings. ModSkill's ability to focus on body-part-level skills makes it a powerful system for controlling complex motions and adapting learned behaviors across different tasks. By utilizing a skill modularization attention layer, ModSkill enhances the generalization of motor skills across various tasks, such as reaching or striking, thereby further improving task-specific adaptation.

\noindent\textbf{From primitive skills to high-level goal execution}. The success of BFMs in learning diverse primitive skills has propelled the development of more advanced BFMs capable of interpreting and executing high-level goals, including language instructions and multi-task objectives. A notable example is MoConVQ \cite{yao2024moconvq}, which introduces a unified motion control framework based on discrete latent codes learned via vector quantized variational autoencoder (VQ-VAE). The model supports a wide range of downstream tasks, including motion tracking, interactive control, and text-to-motion generation, by offering a compact and modular representation. MoConVQ also integrates with large language models (LLMs) and enables the simulated agents to be directed via in-context language prompts, thereby bridging symbolic reasoning and physical control. 

\begin{figure}[h!]
    \centering
    \includegraphics[width=\linewidth]{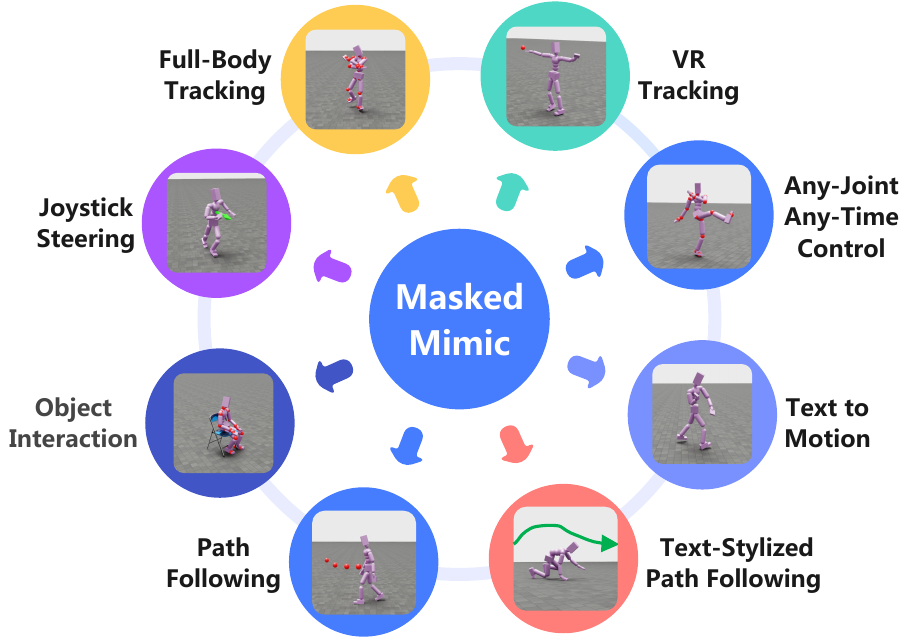}
    \caption{MaskedMimic \cite{tessler2024maskedmimic} is a representative BFM trained via goal-conditioned learning, which supports diverse control modalities and achieves seamless transitions between distinct tasks.}
    \label{fig:maskedmimic}
\end{figure}

Meanwhile, MaskedMimic \cite{tessler2024maskedmimic} addresses physics-based character control as a general motion inpainting problem, producing full-body motions from partial descriptio, such as masked keyframes, objects, or text instructions. MaskedMimic involves a two-phase training process. First, a fully-constrained motion tracking controller is trained to imitate diverse reference motions. Then, a partially-constrained VAE-based policy distills this knowledge through masked goal conditioning. As a result, MaskedMimic can dynamically adapt to complex scenes and support applications ranging from VR control to complex human-object interaction (HOI), as illustrated in Fig.~\ref{fig:maskedmimic}. In addition, InterMimic \cite{xu2025intermimic} focuses on the HOI scenario and designs a two-stage teacher-student framework that distills imperfect motion capture interaction data into robust and physics-based controllers. Teacher policies are trained on subsets of noisy data and refined through simulation, then distilled into a student policy with RL-based fine-tuning. This curriculum strategy enables generalization across diverse interactions with high physical fidelity. 

For real-world robotic applications, HOVER \cite{he2025hover} introduces a multi-mode policy distillation framework that enables humanoid robots to switch seamlessly between tasks (\textit{e.g.}, locomotion, manipulation, and navigation) using a single unified policy distilled from an oracle. This eliminates the need for task-specific controllers, demonstrating general-purpose control in real-world environments, akin to the versatility seen in MaskedMimic for virtual characters.

All the above methods follow the idea of BFM, which is revealed from two aspects: (i) their ability to learn broad behavioral priors from diverse data sources, and (ii) their fast adaptation ability to downstream tasks. Trained on large-scale data, these models generalize across a wide range of behaviors without being limited to a single domain. For example, InterMimic handles various HOI tasks, while MoConVQ adapts to tasks like goal-reaching and text-conditioned motion generation. Critically, they exhibit rapid adaptation to new scenarios with minimal retraining. Thus, the learned broad behavioral priors, along with the swift adaptation capability, characterize these methods as BFMs.

\subsubsection{Intrinsic Reward-driven Learning}

\begin{figure}[t!]
    \centering
    \includegraphics[width=\linewidth]{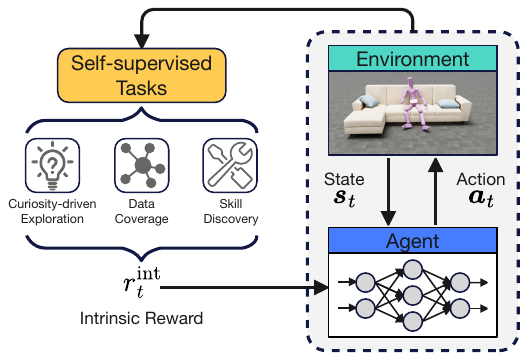}
    \caption{Workflow of the intrinsic reward-driven learning. The agent is trained to explore and comprehend the environment via self-supervised reward signals, thereby achieving non-directional skill acquisition.}
    \label{fig:intrinsic}
\end{figure}

In tracking-based learning, the agent is consistently provided with an explicit objective (\textit{e.g.}, joint angles or velocities) and trained via a well-specified reward function to achieve targeted skill acquisition. In contrast, intrinsic reward-driven learning presents a distinct approach, where the agent is motivated to explore the environment without relying on explicit task-specific rewards. Instead, as illustrated in Fig.~\ref{fig:intrinsic}, the agent is guided by intrinsic rewards, which are self-generated signals that encourage exploration, skill acquisition, or novelty detection. Extensive strategies for intrinsic reward-driven learning have been developed, including curiosity-driven exploration \cite{pathak2017curiosity,burda2018exploration,pathak2019self,sekar2020planning}, skill discovery \cite{gregor2017variational,eysenbach2018diversity,hansen2020fast,liu2021aps}, and maximizing data coverage \cite{bellemare2016unifying,ostrovski2017count,seo2021state,yarats2021reinforcement}, each encouraging the agent to explore different aspects of the environment. Denote by $r^{\rm int}:\mathcal{S}\rightarrow\mathbb{R}$ the intrinsic reward function, the optimization objective in Eq.~(\ref{eq:rl}) is reformulated as
\begin{equation}
    J_{\pi}=\mathbb{E}_{\pi}\left[\sum_{t=0}^{\infty}\gamma^{t}r^{\rm int}(\bm{s}_t)\right].
\end{equation}

For example, the intrinsic curiosity module (ICM) \cite{pathak2017curiosity} encourages the agent to explore unfamiliar states by providing an intrinsic reward based on the prediction error between the current and predicted next state. The intrinsic reward is proportional to the discrepancy between the predicted state and the actual state, motivating the agent to interact with environments that it cannot predict, thereby driving exploration and learning. ICM has been shown to significantly improve the agent's ability to explore complex environments with sparse or no external rewards. In contrast, DIAYN \cite{eysenbach2018diversity} utilizes latent variable discovery to guide the agent's exploration, encouraging it to maximize the diversity of behaviors exhibited. DIAYN introduces an intrinsic reward based on the mutual information between the agent's latent skill variable and its environment state, encouraging the agent to discover and explore a wide range of distinct behaviors. By learning a set of diverse, reusable skills, DIAYN enables agents to tackle complex tasks without requiring domain-specific rewards, making it a valuable approach for unsupervised skill discovery.

In addition, RE3 \cite{seo2021state} focuses on state coverage maximization, where the intrinsic reward is driven by the state visitation frequency. The goal is to encourage the agent to explore states that are infrequent or underrepresented in its past experience. By learning from these underexplored regions, RE3 enhances the agent's understanding of the environment and helps prevent the agent from getting stuck in local optima. RE3 allows the agent to explore a broader range of states, improving the diversity of the learned representations and making it suitable for environments with sparse external rewards. Following DIAYN and RE3, ODPP \cite{chen2023unified} is proposed as a unified framework to discover skills that are both diverse and have superior state coverage, based on a novel use of the determinant point process. Specifically, the unsupervised objective is to maximize (i) the number of modes covered within each trajectory to enhance state coverage, and (ii) the number of modes across the trajectory space to promote skill diversity.

Compared to BFMs trained on large-scale and diverse behavior datasets, intrinsic reward-driven models are typically trained on a single environment/task, which limits their ability to provide strong and intuitive behavioral priors. However, these models exhibit several key properties desirable for BFM pre-training. Specifically, they can (i) effectively explore the environment and action space and discover generalizable and latent behavioral priors under the motivation of intrinsic rewards, and (ii) demonstrate the capability to learn and adapt to various downstream tasks, as shown in \cite{yarats2021reinforcement,laskin2urlb}. Therefore, we consider these methods as influential and conceptually related prototypes, whose unsupervised behavioral prior acquisition mechanisms can inspire the development of more effective BFMs. Future work may combine the exploratory advantages of intrinsic rewards with the task-specific utility of goal-conditioned learning, leading to a more comprehensive approach to behavioral prior learning.

\subsubsection{Forward-backward Representation Learning}
\begin{figure}[t!]
    \centering
    \includegraphics[width=\linewidth]{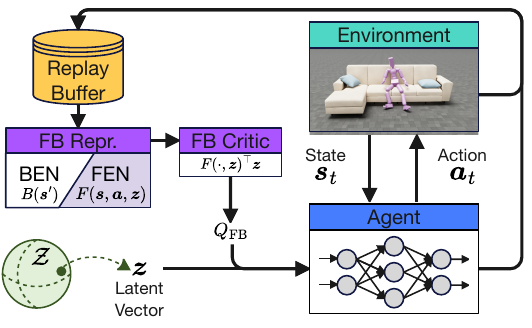}
    \caption{Workflow of the forward-backward representation Learning, where a forward embedding network (FEN) and backward embedding network (BEN) are employed to learn the approximation of the successor measure, thereby achieving the universal policy representation.}
    \label{fig:fb}
\end{figure}

Recent advances in BFM are propelled by a novel framework entitled forward-backward (FB) representation learning \cite{blier2021learning}. As illustrated in Fig.~\ref{fig:fb}, it disentangles policy learning from task-specific objectives, which is fundamentally different from goal-conditioned learning and intrinsic reward-driven learning. By learning a universal policy representation, it can be rapidly adapted to new tasks through reward inference or demonstration alignment, without additional environment interaction or policy optimization.

At its core, the FB representation learning seeks to learn a finite-rank approximation of the successor measure, which is an extension of the successor representation \cite{dayan1993improving,kulkarni2016deep}. It depicts the discounted future state visitation distribution as a measure over states. For each policy $\pi$, its successor measure is defined as

\begin{equation}
    M^{\pi}(\mathcal{X}| \bm{s}, \bm{a})\coloneqq\sum_{t=0}^{\infty} \gamma^{t} \operatorname{Pr}\left(\bm{s}_{t+1} \in \mathcal{X} | \bm{s}, \bm{a}, \pi\right), \forall \mathcal{X} \subset \mathcal{S}.
\end{equation}
The successor measure satisfies a measure-valued Bellman equation \cite{blier2021learning}:
\begin{equation}
\begin{aligned}
M^{\pi}(\mathcal{X}| \bm{s}, \bm{a})&=P(\mathcal{X}|\bm{s}, \bm{a}) \\
+\gamma \mathbb{E}_{\bm{s}' \sim P(\cdot | \bm{s}, \bm{a}), \bm{a}' \sim \pi(\cdot | \bm{s}')}&\left[M^{\pi}\left(\mathcal{X} | \bm{s}', \bm{a}'\right)\right], \mathcal{X} \subset \mathcal{S}.
\end{aligned}
\end{equation}

Equipped with the successor measure, the action-value function $Q^{\pi}_r(\bm{s},\bm{a})$ of $\pi$ for any reward function $r$ satisfies
\begin{equation}\label{eq:avf sm}
\begin{aligned}
    Q_{r}^{\pi}(\bm{s}, \bm{a})\coloneqq&\mathbb{E}_{\pi}\left[\sum_{t=0}^{\infty} \gamma^{t} r\left(\bm{s}_{t+1}\right) | \bm{s}, \bm{a}, \pi\right]\\
    =&\int_{\bm{s}' \in \mathcal{S}} M^{\pi}\left(\mathrm{d} \bm{s}' | \bm{s}, \bm{a}\right) r\left(\bm{s}'\right).
\end{aligned}
\end{equation}
Eq.~(\ref{eq:avf sm}) decouples the action-value function as two separate terms: (i) the successor
measure that models the evolution of the policy in the environment, and (ii) the reward function that captures task-relevant information. This factorization suggests that learning the successor measure
for $\pi$ allows for the zero-shot evaluation of $Q^{\pi}_r$ on any reward without further training. 

Notably, the success measure can be estimated as \cite{touati2021learning}
\begin{equation}
    M^{\pi}(\mathcal{X}| \bm{s}, \bm{a})\approx\int_{\bm{s}' \in \mathcal{X}}F(\bm{s},\bm{a})^\top B(\bm{s}')\rho(\mathrm{d}\bm{s}'),
\end{equation}
where $\rho$ is an arbitrary distribution over states, $F:\mathcal{S}\times\mathcal{A}\rightarrow \mathbb{R}^{d}$ is the forward embedding and $B:\mathcal{S}\rightarrow \mathbb{R}^{d}$ is the backward embedding, respectively. Denote by $\bm{z}=\mathbb{E}_{\bm{s}\sim\rho}\left[B(\bm{s})r(\bm{s})\right]$, the action-value function is rewritten as 
\begin{equation}
    Q^{\pi}_{r}=F(\bm{s},\bm{a})^\top \bm{z}.
\end{equation}

To learn a family of polices, both the forward embedding $F$ and policy $\pi$ can be parameterized by the same task encoding vector $\bm{z}$ \cite{touati2021learning}, such that
\begin{equation}
    M^{\pi_{\bm{z}}}(\mathcal{X}| \bm{s}, \bm{a})\approx\int_{\bm{s}' \in \mathcal{X}}F(\bm{s},\bm{a},\bm{z})^\top B(\bm{s}')\rho(\mathrm{d}\bm{s}'),
\end{equation}
where $\bm{z}\subseteq \mathbb{R}^{d}$, and the policy $\pi_{\bm z}$ is defined as
\begin{equation}
    \pi_{\bm z}=\underset{\bm{a}}{\rm argmax}\:F(\bm{s},\bm{a},\bm{z})^\top\bm{z}.
\end{equation}
Then, the forward-backward embedding network is trained to minimize the temporal difference (TD) loss derived as the Bellman residual:
\begin{equation}
\begin{aligned} 
L_{\rm FB}&=  \mathbb{E}_{\substack{z \sim \nu, (\bm{s}, \bm{a}, \bm{s}') \sim \rho, \\ \bm{s}^{+} \sim \rho, \bm{a}' \sim \pi_{\bm z}(\bm{s}')}} \bigg[ \big( F(\bm{s}, \bm{a}, \bm{z})^{\top} B(\bm{s}^{+}) \\
& - \gamma\cdot\mathrm{sg}(F)(\bm{s}', \bm{a}', \bm{z})^{\top}\mathrm{sg}(B)(\bm{s}^{+}) \big)^2 \bigg] \\ 
& - 2 \mathbb{E}_{\bm{z} \sim \nu, (\bm{s}, \bm{a}, \bm{s}') \sim \rho} \left[ F(\bm{s}, \bm{a}, \bm{z})^{\top} B(\bm{s}') \right], 
\end{aligned}
\end{equation}
where $\bm{s}^{+}$ denotes a future state and ${\rm sg}(\cdot)$ denote the stop-gradient operation. Consider continuous action spaces, the policies can be obtained by training an actor network to minimize
\begin{equation}
    L_{\rm actor}=-\mathbb{E}_{\substack{z \sim \nu, \bm{s} \sim \rho, \bm{a} \sim \pi_{\bm z}(\bm{s})}} \left[ F(\bm{s}, \bm{a}, \bm{z})^{\top} \bm{z}\right].
\end{equation}

Once the FB model is trained, it can be utilized to solve diverse tasks in a zero-shot manner without performing additional task-specific learning, planning, or fine-tuning. For example, given a task reward function $r$, the policy can be inferred by computing
\begin{equation}
    \bm{z}_r=\frac{1}{n}\sum_{i=1}^{n}r(\bm{s}_i)B(\bm{s}_i),
\end{equation}
where $\{\bm{s}_i\}_{i=1}^{n}$ is a set of sample states. Similarly, for a goal-reaching problem, it suffices to compute the encoder vector by $\bm{z}_{\bm{s}}=B(\bm{s}),\bm{s}\in\mathcal{S}$.

As introduced in Section~\ref{sec:intro-bfm}, the term "BFM" is first literally introduced in the FB-IL framework \cite{pirotta2024fast}, which supports multiple IL principles without needing separate RL routines for each new task. FB-AW \cite{cetin2025finer} enhances the FB framework by incorporating the advantage-weighted (AW) regression loss in offline RL algorithms, achieving stable performance on complex datasets (\textit{e.g.}, DMC Humanoid \cite{tassa2018deepmind}) composed of near-optimal trajectories. The subsequent FB-AWARE \cite{cetin2025finer} method further employs an auto-regressive encoding (ARE) mechanism to let fine-grained task features depend on coarser-grained task information. While the standard FB method employs a linear task projection that can blur rewards and reduce spatial precision, this mechanism can significantly enhance the expressivity and performance of the FB framework, particularly for tasks that require spatial accuracy or generalization. Notably, the AW and ARE mechanisms can be flexibly combined based on the task requirements. For example, the ARE mechanism can be used independently in the FB-ARE variant for environments where the AW component is not required for satisfactory performance.

\begin{figure}[t!]
    \centering
    \includegraphics[width=\linewidth]{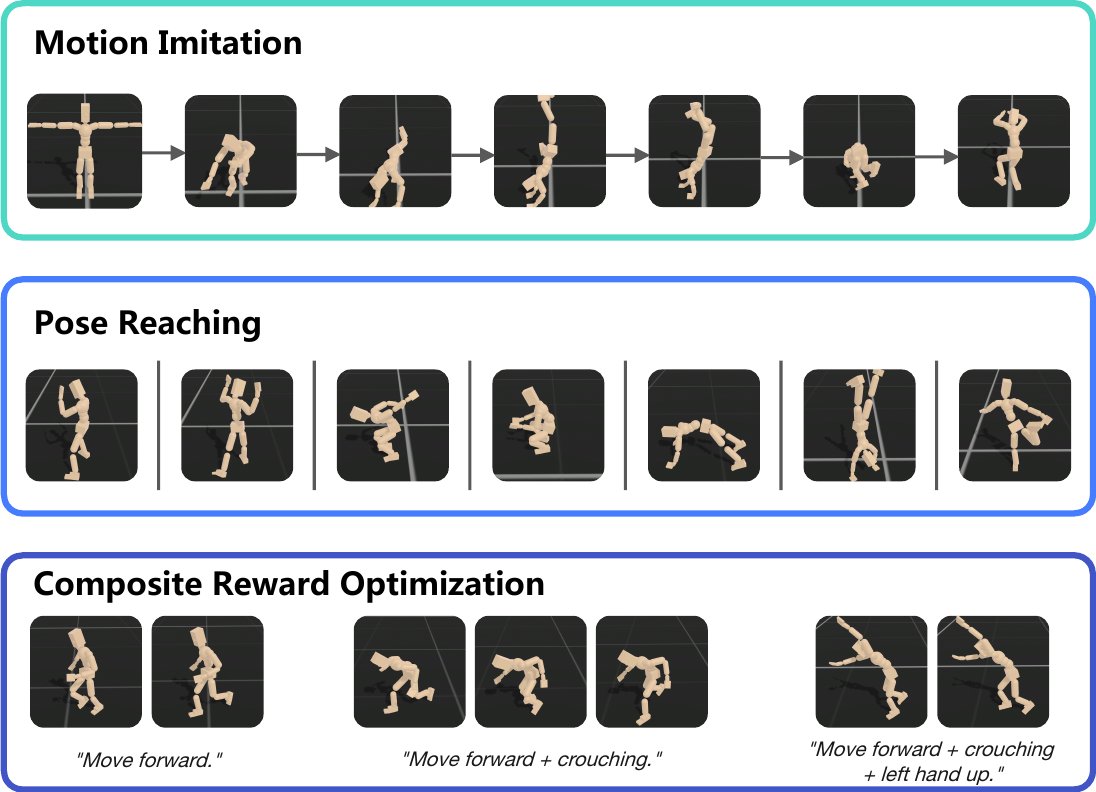}
    \caption{Motivo \cite{tirinzoni2025zeroshot} learns broad behavior priors and demonstrates outstanding zero-shot adaptation capability to diverse downstream tasks, including complex motion imitation, pose reaching, and composite reward optimization. Moreover, Motivo achieves real-time motor control while ensuring motion naturalness.}
    \label{fig:motivo}
\end{figure}

The FB framework provides a general and flexible approach for training BFMs by learning successor measure representations and applying pre-trained policies to new tasks. However, it suffers from several limitations: (i) when the latent dimension $d$ is finite, it relies on the assumption of a low-rank dynamic, leading to a limited inductive bias for policy selection; (ii) poor coverage in the training dataset causes offline learning to fail in reliably optimizing policies, often collapsing to a few suboptimal behaviors with weak performance on downstream tasks. These limitations greatly hinder the application of the FB framework for humanoid robots. To address these limitations, the Motivo framework \cite{tirinzoni2025zeroshot} combines the FB framework with a conditional policy regularization (FB-CPR) term. As illustrated in Fig.~\ref{fig:motivo}, Motivo solves diverse humanoid WBC tasks in a zero-shot manner, including motion tracking, goal reaching, and reward optimization.

Specifically, FB-CPR learns the FB representations with a discriminator-based regularization scheme, whose loss function is defined as
\begin{equation}
\begin{aligned}
    L_{\rm FB-CPR}=&-\mathbb{E}_{\substack{\bm{z} \sim \nu,\\ \bm{s} \sim \mathcal{D}_{\rm online}, \bm{a} \sim \pi_{\bm z}(\cdot | \bm{s})}}\left[F(\bm{s}, \bm{a}, \bm{z})^{\top} \bm{z}\right]\\
    &+\alpha \mathrm{KL}\left(p_{\pi}, p_{\mathcal{E}}\right),
\end{aligned}
\end{equation}
where $\mathcal{D}_{\rm online}$ is the associated replay buffer of unsupervised transitions, $p_{\pi}(\bm{s},\bm{z})$ is the joint distribution of $(\bm{s},\bm{z})$ induced by FB, and $p_{\mathcal{E}}$ is the joint distribution of the dataset. However, it is intractable to optimize the divergence term directly via a RL procedure. To tackle the problem, FB-CPR interprets the divergence as an expected return under the policies and defines a divergence-based reward $r^{\rm div}$ as
\begin{equation}
\begin{aligned}
&\mathrm{KL}\left(p_{\pi}, p_{\mathcal{E}}\right)=\mathbb{E}_{\substack{\bm{z} \sim \nu, \\ \bm{s} \sim \rho^{\pi}_{\bm{z}}}}\left[\log \frac{p_{\pi}(\bm{s}, \bm{z})}{p_{\mathcal{E}}(\bm{s}, \bm{z})}\right]\\
=-\mathbb{E}_{\bm{z} \sim \nu} &\mathbb{E}\left[\sum_{t=0}^{\infty} \gamma^{t} \log \frac{p_{\mathcal{E}}\left(\bm{s}_{t+1}, \bm{z}\right)}{p_{\pi}\left(\bm{s}_{t+1}, \bm{z}\right)} \,\middle|\, \bm{s}_0 \sim \mu, \pi_{\bm{z}}\right].
\end{aligned}
\end{equation}
Then, a discriminator network $D: \mathcal{S}\times\mathcal{A}\rightarrow[0,1]$ is trained to estimate the $r^{\rm div}$:
\begin{equation}
    r^{\rm div}=\log \frac{p_{\pi}(\bm{s}, \bm{z})}{p_{\mathcal{E}}(\bm{s}, \bm{z})}\approx\log \frac{D}{1-D}.
\end{equation}
The estimation holds due to the optimal discriminator satisfies $D^*=\frac{p_{\mathcal{E}}}{p_{\mathcal{E}}+p_{\pi}}$ \cite{goodfellow2014@generative}. Finally, the divergence term can be estimated by training another critic network via off-policy TD learning, and the actor loss for FB-CPR is rewritten as
\begin{equation}
\begin{aligned}
    L_{\rm FB-CPR}=&-\mathbb{E}_{\substack{\bm{z} \sim \nu,\\ \bm{s} \sim \mathcal{D}_{\rm online}, \bm{a} \sim \pi_{\bm z}(\cdot | \bm{s})}}\left[F(\bm{s}, \bm{a}, \bm{z})^{\top} \bm{z}\right]\\
    &+\alpha Q_{r^{\rm div}}^{\pi}(\bm{s}, \bm{a}, \bm{z}).
\end{aligned}
\end{equation}

It is natural to find that FB-CPR is not a rigorous unsupervised method, as it leverages unlabeled demo data to assist motion prior learning. By aligning unsupervised RL with human-like behavioral priors from unlabeled data, FB-CPR enhances policy diversity and dataset coverage, enabling the agent to learn a rich latent space of behaviors (\textit{e.g.}, walking, jumping, handstands) and achieve robust zero-shot performance across diverse tasks. Experimental results demonstrate that Motivo achieves an 83\% success rate in motion tracking tasks, and performs 61\% of the top-line performance in reward optimization tasks, surpassing DIFFUSER \cite{janner2022planning} in computational efficiency by requiring only 12 seconds per 300-step episode. Additionally, it outperforms ASE and CALM in motion diversity, achieving a score of 4.70 ($\pm$0.66), reflecting its ability to capture a broader range of behaviors. 

Moving beyond simulated environments, BFM-Zero \cite{li2025bfm} first successfully deploys BFMs derived from the FB framework in real-world humanoid robots by introducing key architectural innovations that enhance sim-to-real robustness. Building upon the FB-CPR foundation, BFM-Zero incorporates a dual-critic architecture that augments the adversarial imitation objective with a separate critic network explicitly trained to maximize auxiliary stability rewards, such as action smoothness and joint limit avoidance. Meanwhile, BFM-zero integrates extensive domain randomization, history-based policies, and a scaled-up transformer-inspired residual network. These efforts collectively enable the learned policies to exhibit real-time, stable, and naturalistic WBC directly transferred from simulation to hardware, such as the Unitree G1 and Booster T1 robots, without any fine-tuning. 

\subsection{Adaptation}
Equipped with the derived BFMs through the aforementioned pre-training frameworks, we next introduce recent advancements in the adaptation techniques of BFMs. These approaches can be broadly categorized into two types: \textbf{fine-tuning} and \textbf{towards hierarchical control}. It is worthwhile to highlight the distinction between the adaptation capabilities arising directly from pre-training and the fine-tuning approaches presented in this section. Pre-trained BFMs often exhibit representation-level generalization, allowing them to solve downstream tasks directly through task conditioning, such as providing goal states, latent vectors, or reward embeddings, without updating model parameters. For example, Motivo \cite{tirinzoni2025zeroshot} can generate desirable locomotion simply by constructing a reward function that specifies the intended behavior. In contrast, the adaptation techniques reviewed below often involve refining parameters or introducing additional modules to further enhance performance in specific tasks or environments.

\subsubsection{Fine-tuning}

Fine-tuning of BFMs seeks to bridge the gap between general-purpose behavioral priors and task-specific requirements. While pre-trained BFMs capture broad behavioral priors, they often lack precision for specialized tasks or novel environments. To address this, fast adaptation techniques such as residual latent adaptation (ReLA) and lookahead latent adaptation (LoLA) \cite{sikchi2025fast} enable BFMs to achieve zero-shot adaptation with minimal online interactions after pre-training. These methods improve task performance by up to 40\%, demonstrating the effectiveness of fast adaptation strategies that enable efficient task switching without requiring extensive retraining. In contrast, methods like TaskTokens \cite{vainshtein2025task} enhance goal-conditioned BFMs by generating task-specific tokens through a task encoder, achieving high success rates across complex tasks such as motion tracking and goal-reaching. TaskTokens shows up to 99.75\% success in tasks such as the long jump. Furthermore, the belief-FB method \cite{bobrin2025zero} extends forward–backward representations by incorporating a transformer-based dynamics inference module, substantially improving robustness to unseen environmental variations and achieving a performance improvement of up to $2\times$. Future work could investigate additional post-training techniques, such as test-time scaling \cite{gandhi2024stream} or RL from human feedback \cite{ouyang2022training}, to further improve the adaptability and efficiency of BFMs and ensure alignment with human preferences in real-world applications.

\subsubsection{Towards Hierarchical Control}
While fine-tuning techniques enhance the performance of BFMs for specific tasks through minimal modifications during test time, several pioneering works have attempted to establish a hierarchical control architecture based on BFM, which decouples high-level planning from low-level motion execution to achieve more scalable and flexible control \cite{shao2025langwbc,xue2025leverb}. For example, UniHSI \cite{xiao2024unified} introduces a unified framework for human-scene interaction by using language commands to guide a chain-of-contacts, which represents the sequence of human-object contact pairs. The system translates language inputs into structured task plans, which are then executed by a unified controller based on the AMP architecture. This framework achieves semantic alignment between language commands and physical motions, supporting diverse interactions with single or multiple objects. TokenHSI \cite{pan2025tokenhsi} further extends this line by proposing a transformer-based unified policy that tokenizes human proprioception and task states. By separating shared motor knowledge (proprioception token) from task-specific parameters (task tokens), it enables seamless multi-skill unification and flexible adaptation to novel tasks, such as skill composition (\textit{e.g.}, carrying while sitting), variations in object or terrain shape, and long-horizon task completion. 

In contrast, CLoSD \cite{tevet2025closing} introduces a text-driven RL controller that combines motion diffusion models with physics-based simulations for robust multi-task control of human characters. By utilizing a real-time diffusion planner and a motion tracking controller in a closed-loop feedback system, CLoSD can handle complex tasks, such as goal-reaching, striking, and human-object interactions, all of which are controlled through text prompts and target locations. Similarly, UniPhys \cite{wu2025uniphys} introduces a diffusion-based behavior cloning framework that unifies planning and control with diffusion forcing to handle prediction errors, enabling flexible control via text, velocity, and goal guidance for applications, such as dynamic obstacle avoidance and long-horizon planning. These applications demonstrate that BFMs serve as a pivotal bridge between high-level semantic instructions and low-level physical execution, leveraging pre-trained behavioral priors to enable zero-shot adaptation, multi-task generalization, and physics-aware motion synthesis. By integrating various control paradigms (\textit{e.g.}, transformer-based tokenization and closed-loop diffusion planning), these highlight the potential of BFMs to democratize humanoid control across complex, real-world scenarios, ranging from interactive robotics to dynamic environment adaptation.

\section{Applications and Limitations}\label{sec:bfm app limit}
\begin{figure}[h!]
    \centering
    \includegraphics[width=\linewidth]{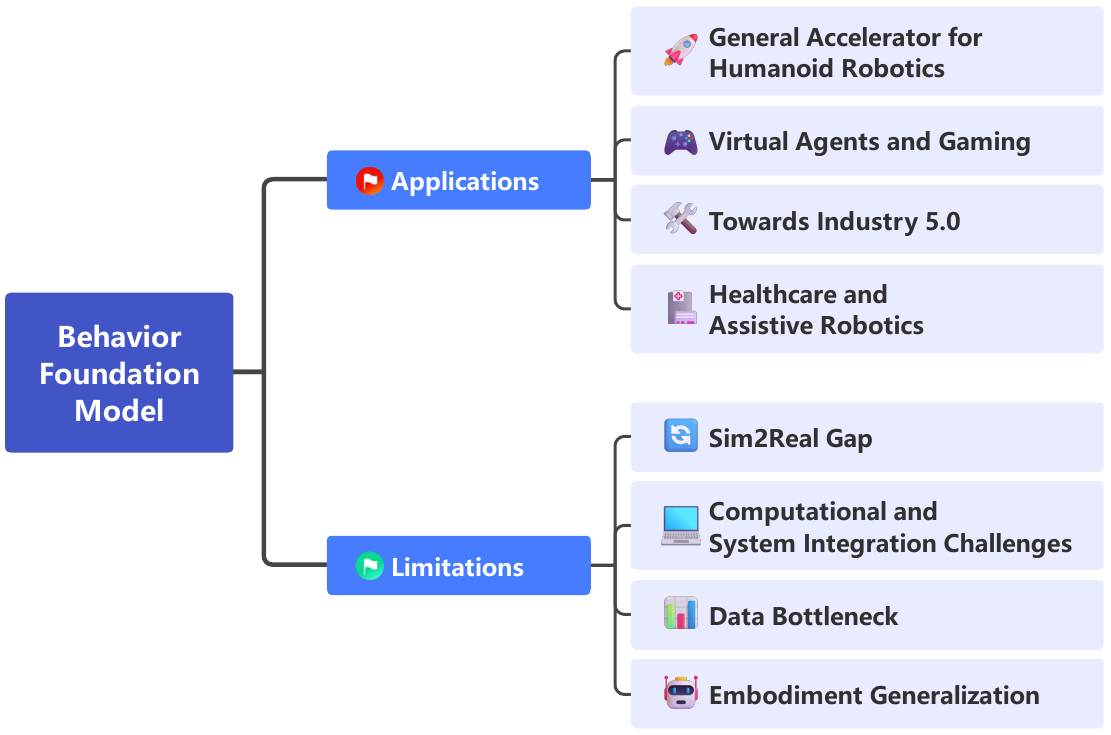}
    \caption{An overview of the practical applications and current limitations of BFMs.}
    \label{fig:app_limits}
\end{figure}

BFMs are foreseen to significantly enhance humanoid robotics by providing a universal pre-trained controller capable of generalizing across diverse tasks. In this section, we explore the potential applications of BFMs in various industries, including healthcare robotics and gaming. Furthermore, we identify the key limitations of current BFMs, including the Sim2Real gap, data bottleneck, and generalization to embodiment. An overview of the applications and limitations is illustrated in Fig.~\ref{fig:app_limits}. Our analysis is inspired by the current development and applications of other foundation models, such as LLMs \cite{naveed2025comprehensive} and large vision models \cite{liu2024sora,awais2025foundation}.

\subsection{Applications}
\subsubsection{General Accelerator for Humanoid Robotics}
BFMs will act as a transformative general accelerator for humanoid robotics, accelerating the development and deployment of advanced WBC systems. Unlike the previous pipelines that require resource-intensive and task-specific training, BFMs eliminate the need for training from scratch by pre-training on vast and diverse behavior datasets, thereby embedding a rich behavioral prior that accelerates downstream adaptation exponentially \cite{sikchi2025fast}. Moreover, advanced BFMs with zero-shot adaptation capabilities, such as Motivo \cite{tirinzoni2025zeroshot}, can directly map high-level task specifications (\textit{e.g.}, goal states and reward functions) to low-level control actions, bypassing traditional RL loops entirely. This capability efficiently solves basic control tasks but also facilitates rapid prototyping, allowing developers to evaluate robot behaviors in both simulation and real-world environments within minutes, dramatically shortening the development cycle.

\subsubsection{Virtual Agents and Gaming}
Virtual agents play a crucial role in advancing the development of BFMs for humanoid robots. Many foundational BFM techniques (\textit{e.g.}, large-scale motion pre-training and latent skill learning) are first developed and validated in simulation environments for virtual agents before being adapted for real-world humanoids. On the other hand, the strong growth of the game industry and metaverse platforms creates an increasing demand for natural, physically grounded avatar motions, which align closely with the objectives of humanoid BFMs \cite{kopel2018implementing,mehta2022exploring,uludaugli2023non}. BFMs offer a groundbreaking solution by enabling lifelike, context-aware non-player characters (NPCs) behaviors without extensive manual scripting. Pre-trained on diverse human behavior datasets, BFMs generate adaptive actions, such as tactical combat, social engagement, or exploration, seamlessly responding to dynamic player inputs. By integrating BFMs with LLMs, NPCs can interpret complex player instructions (\textit{e.g.}, dialogue-driven commands in role-playing games) and foster immersive and responsive interactions. This capability positions BFMs as a pivotal technology for revolutionizing virtual agents, enabling next-generation gaming experiences with unprecedented behavioral realism and interactivity.

\subsubsection{Towards Industry 5.0}

While Industry 4.0 introduced smart factories with cyber-physical systems, IoT, and AI-driven automation, Industry 5.0 shifts toward human-centric, resilient, and sustainable manufacturing, emphasizing collaborative robotics, adaptive intelligence, and personalized production \cite{xu2021industry,leng2022industry,huang2022industry,akundi2022state,hassan2024systematic}. To that end, robots must move beyond rigid automation and instead exhibit generalizable, adaptive, and explainable behaviors \cite{dzedzickis2021advanced,bartovs2021overview,arents2022smart,soori2024intelligent,lee2022does}. BFMs are expected to empower this landscape by enabling humanoid robots to seamlessly blend pre-trained motor skills with real-time adaptability, effortlessly switching between tasks like precision welding and adaptive part handling. By integrating large multimodal models with BFMs, robots can process diverse inputs like gestures, voice commands (\textit{e.g.}, "handle gently"), or environmental cues, fostering intuitive human-robot collaboration in shared workspaces. BFMs also ensure resilience, autonomously recovering from disturbances like unbalanced loads in logistics, and support personalized production through zero-shot or few-shot learning.

\subsubsection{Healthcare and Assistive Robotics}

The global population aging presents unprecedented challenges for healthcare systems \cite{rowland2009global, bloom2015global, navaneetham2023global, zhang2025impact}, resulting in an increased demand for assistive technologies that support independent living and rehabilitation. Extensive and diverse robots have been developed for robot-assisted dressing, rehabilitation therapy, medical treatment, and caregiving for the elderly and children \cite{feil2005defining,miller2006assistive,okamura2010medical,riek2017healthcare,holland2021service,kyrarini2021survey,sanchez2021textile,zhang2022learning,javaid2025utilization}. Humanoid robots are ideally suited for these tasks due to their anthropomorphic design, navigating human-centric environments and perform precise, natural, and intuitive interactions. BFMs offer a promising solution by enabling robots to adapt to diverse user needs and unstructured environments. For instance, BFMs can empower assistive robots to perform tasks like mobility support (\textit{e.g.}, fall prevention, gait assistance) or daily tasks (\textit{e.g.}, object retrieval, meal preparation) with minimal user-specific tuning. In rehabilitation, BFMs trained on clinician-guided demonstrations can personalize therapy protocols by dynamically adjusting task difficulty or providing real-time feedback based on patient progress.

\subsection{Limitations}
\subsubsection{Sim2Real Gap} \label{sec:sec:sim2real}
The Sim2Real gap is a persistent challenge in robotics, representing the performance discrepancy between the policies trained in simulators and their real-world deployment \cite{kadian2020sim2real,hofer2021sim2real,iyengar2023sim2real,su2024sim2real}. Traditional model-based controllers address this through explicit physics modeling and robust optimization, while data-driven approaches employ domain randomization and system identification \cite{kaspar2020sim2real,horvath2022object,huber2024domain,yao2025sim2real}. Recent advancements like ASAP \cite{he2025asap} mitigate dynamics mismatch via residual action learning yet face policy-specific limitations, as their residuals are trained on trajectories from a single pre-trained policy, thereby restricting generalization. BFMs exacerbate this challenge by encoding a vast spectrum of behaviors, ranging from locomotion to multi-contact interactions, which introduces high-dimensional transfer risks. For example, a BFM trained for diverse humanoid motions may fail to adapt to real-world actuator delays, friction variations, or sensor noise, resulting in unstable or unsafe execution. The integration of visual signals into control systems further compounds the challenge, and perceptual domain shifts (\textit{e.g.}, lighting, texture, or camera calibration mismatches) and generalization gaps in visual features can destabilize motion policies based on simulated visual inputs. Current BFMs remain largely confined to simulation, with no documented large-scale real-world deployments. While several pioneering works have been devoted to developing BFM-like controllers in real humanoid robots, such as HugWBC \cite{xue2025unified} and CLONE \cite{li2025clone}, their motion skills remain limited, highlighting a significant challenge to Sim2Real feasibility in maintaining behavior richness. This gap stems from behavioral overgeneralization, dynamics mismatches, and latent space instability, which hinder scaling targeted Sim2Real successes (\textit{e.g.}, quadruped locomotion or grasping) to BFM's complexity.

\subsubsection{Computational and System Integration Challenges}
While the Sim2Real gap presents deployment challenges at the data and policy levels, another significant challenge in deploying BFMs arises from computational and system integration issues. To ensure responsive motion execution and effective disturbance rejection, the humanoid control policy typically requires real-time command updates at frequencies ranging from 50Hz to 100Hz \cite{li2025bfm, hwangbo2017control}. Unlike simulation environments, where computing resources are plentiful, real-world humanoids operate under strict constraints related to latency, power consumption, and thermal management. Although BFMs can excel with increased model size, as demonstrated by Motivo \cite{tirinzoni2025zeroshot}, these constraints limit the feasibility of using larger models. For instance, the BFM-Zero \cite{li2025bfm} only has a maximum parameter count of 44.5M, while SONIC \cite{luo2025sonic} has 42M parameters.

On the other hand, incorporating perceptual input (\textit{e.g.}, RGB or depth images) into BFMs further exacerbates the computational challenge, often requiring additional hardware acceleration to meet real-time processing demands. Devices like the NVIDIA Jetson computing board~\cite{biddulph2018comparing, song2025gait} are commonly employed to offload the heavy computational burden of vision-based inference, enabling the robot to process complex multimodal data. However, integrating such hardware introduces significant complexity, particularly in synchronizing the robot's control system with the inference hardware, managing data communication, and minimizing latency. The increased complexity can complicate overall system coordination, making it difficult to achieve seamless integration between perception, decision-making, and action execution. Future research could address these issues by exploring more efficient sensor fusion methods \cite{alatise2020review}, model compression techniques \cite{choudhary2020comprehensive}, and hybrid inference architectures \cite{pujol2021fog}, etc.

\subsubsection{Data Bottleneck}\label{sec:data bottleneck}
The data bottleneck poses a fundamental constraint in developing BFMs for humanoid robots. While the datasets listed in Table~\ref{tb:datasets} have been successfully employed to train the current BFMs, their scale remains significantly smaller than the datasets used to train LLMs or large vision models. This scarcity is exacerbated when retargeting motions to specific robotic platforms \cite{gleicher1998retargetting}, where subtle morphological differences can incur severe policy performance loss. Real-world robot data is even more constrained due to hardware limitations and safety concerns. The challenge compounds when considering multimodal data requirements. 

\begin{table}[h!]
\centering
\caption{Available humanoid motion datasets for training BFMs.}
\label{tb:datasets}
\begin{tabular}{c|ccc}
\toprule
\textbf{Dataset}                       & \textbf{Clip} & \textbf{Hour} & \textbf{Date} \\
\midrule
KIT-ML \cite{plappert2016kit}          & 3911          & 11.2          & 2016          \\
AMASS \cite{mahmood2019amass}          & 11265         & 40.0          & 2019          \\
LAFAN \cite{harvey2020robust}          & 77            & 4.6           & 2020          \\
BABEL \cite{punnakkal2021babel}        & 13220         & 43.5          & 2021          \\
Posescript \cite{delmas2022posescript} & -             & -             & 2022          \\
HumanML3D \cite{Guo_2022_CVPR}         & 14616         & 28.6          & 2022          \\
Motion-X \cite{lin2023motion}          & 81084         & 144.2         & 2023          \\
Motion-X++ \cite{zhang2025motion}      & 120462        & 180.9         & 2025    \\
PHUMA \cite{lee2025phuma}              & 76000         & 73.0          & 2025 \\
Humanoid-X \cite{mao2025universal}     & 163800        & 240.0         & 2025 \\
\bottomrule
\end{tabular}
\end{table}

Current BFMs predominantly rely on proprioceptive inputs, and integrating exteroceptive sensing for real-world deployment introduces new bottlenecks. Pioneering work, such as OmniH2O \cite{he2024omniho} explores multimodal data collection through teleoperation, pairing head-mounted RGBD cameras with whole-body proprioception and motor commands. While this enables tasks like vision-based autonomy and imitation learning, its 40-minute dataset (OmniH2O-6) remains limited to lab environments and lacks diverse contact dynamics. Currently, there is no existing dataset that provides large-scale and temporally aligned recordings of proprioception, vision, and contact dynamics across diverse environments. Therefore, larger-scale, high-quality, and more curated datasets are urgently needed to enhance the effectiveness of current BFMs and to aid in the development of future models.

\subsubsection{Embodiment Generalization}
Despite the use of the term "BFM", current BFMs are typically trained on specific humanoid robot embodiments with fixed morphology, actuator dynamics, and sensor configurations. While they excel at controlling the robot they were trained on, it is currently intractable for them to generalize to novel embodiments with different body shapes, actuator types, or degrees of freedom. This limitation stems from several key challenges in embodiment generalization. One major issue is morphological mismatch, where policies designed for a specific kinematic or dynamic structure struggle to adapt to robots with differing link lengths, joint types, or mass distributions. Additionally, differences in actuator dynamics, such as variations in torque, latency, or control modes (\textit{e.g.}, position vs. torque control), can destabilize policies when transferred between systems. Sensor diversity introduces another layer of complexity, as BFMs often assume consistent sensor inputs and struggle to manage situations involving missing or additional sensing modalities. Furthermore, reward functions created for one type of embodiment (\textit{e.g.}, humanoid walking) may not effectively translate to others (\textit{e.g.}, hexapod gait patterns), resulting in confusion during task adaptation. These challenges underscore the necessity for more adaptable BFM architectures that can abstract skills across various robotic platforms.

\section{Opportunities and Risks}\label{sec:bfm oppo risks}
\begin{figure}[h!]
    \centering
    \includegraphics[width=\linewidth]{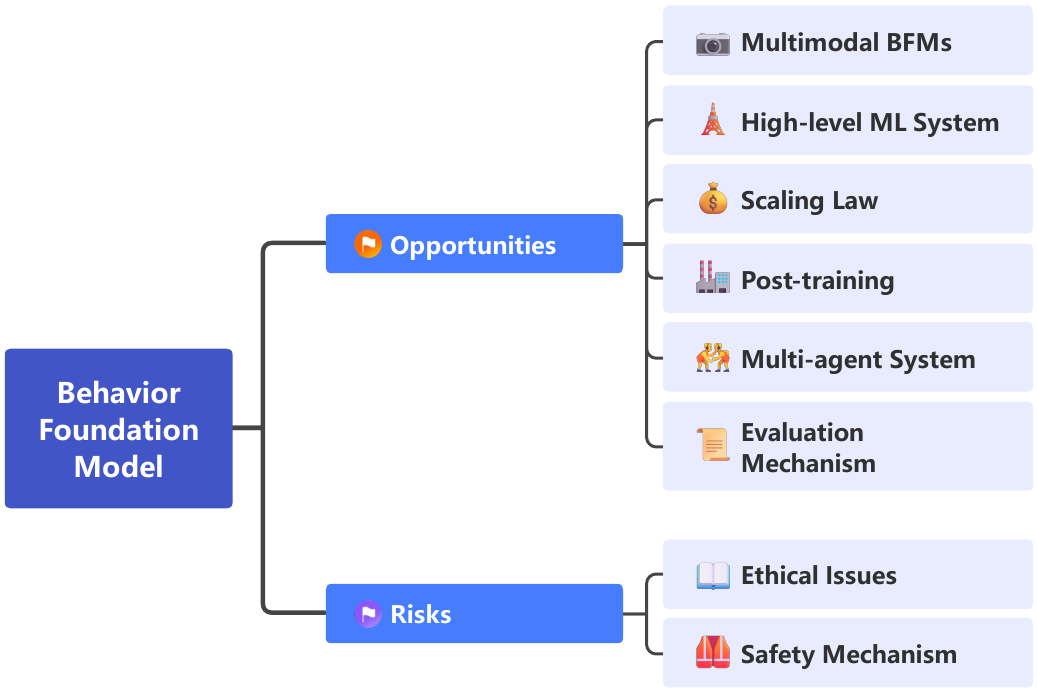}
    \caption{An overview of the future research opportunities and potential risks of BFMs.}
    \label{fig:oppos_risks}
\end{figure}

In this section, we explore key research opportunities that could enhance the broader adoption of BFMs, including their integration into high-level ML systems, scaling laws, post-training techniques, and multi-agent coordination. We also critically examine the risks associated with BFMs, such as ethical concerns in human-robot interaction and safety mechanisms for real-world deployment. An overview of the opportunities and risks is illustrated in Fig.~\ref{fig:oppos_risks}. By addressing these opportunities and risks, the robotics community can ensure that BFMs evolve into robust, adaptable, and socially responsible controllers for next-generation humanoid systems.

\subsection{Opportunities}
\subsubsection{Multimodal BFMs}
A promising direction for BFMs lies in expanding beyond proprioceptive inputs to multimodal sensory integration, incorporating exteroceptive signals such as vision, acoustic signals, and tactile feedback \cite{liu2022multimodal,yao2024multimodal,wang2024multimodal}. While current BFMs primarily rely on proprioceptive information, richer perceptual inputs could enable more robust and adaptive behaviors in unstructured environments. For example, integrating real-time visual perception could enable humanoids to dynamically adjust their movements based on object positions, terrain conditions, or human interactions, enhancing both safety and task performance. However, achieving multimodal BFMs requires the advancement of both large-scale datasets and scalable training paradigms, which are key challenges for the field.

\subsubsection{High-level ML System}
Recent advances in foundation models like LLMs and VLAs have demonstrated their potential as controllers for coordinating specialized AI systems \cite{shen2023hugginggpt,dai2023llmintheloop,chan2024chateval,maeureka,liu2025aligning,yu2025prophet,du2025reinforcement}. For example, HuggingGPT \cite{shen2023hugginggpt} leverages LLMs to orchestrate task planning across diverse models, while Eureka \cite{maeureka} employs LLMs to automate reward function design in RL. These projects present a natural pathway for integrating BFMs into high-level ML systems, where they could serve as universal low-level controllers for humanoid robots. By combining the reasoning capabilities of LLMs (for task planning and adaptation) with the motor priors encoded in BFMs (for real-time execution), such systems could achieve unprecedented flexibility in handling complex and multi-step physical tasks, while minimizing task-specific engineering. The ultimate vision is to create a unified cognitive-physical architecture based on LLMs and BFMs, mirroring the seamless integration of human cognition and physical control.

\subsubsection{Scaling Law}\label{sec:sec:scaling law}
The concept of scaling laws describes how the performance of neural language models improves with increases in model size, data, and compute resources. While these laws are well-established in domains like language and vision, their applicability to BFMs remains an open problem \cite{kaplan2020scaling,clark2022unified,aghajanyan2023scaling,isik2024scaling,que2024d}. Preliminary evidence suggests that scaling BFMs through larger architectures, diverse training datasets, and expanded computational resources can enhance their generalization and zero-shot adaptation capabilities. For instance, FB-CPR \cite{tirinzoni2025zeroshot} demonstrated improved performance in humanoid control tasks with more parameters (from 25M to 288M), achieving more robust zero-shot motion tracking and reward optimization. On the other hand, data scaling appears more critical for BFMs, as it directly determines the quality and physical plausibility of the learned behavior priors, the foundational requirements that model scaling alone cannot address. However, the effect of data scaling remains underexplored, with open questions as discussed in Section~\ref{sec:data bottleneck}. Furthermore, unlike LLMs, BFM scaling must strike an appropriate balance between behavior coverage (diverse motor skills) and control efficiency (precision and real-time stability). Future work should rigorously quantify these scaling dynamics to unlock the full potential of BFMs as general-purpose controllers.

\subsubsection{Post-training}
Post-training techniques have emerged as critical tools for the continued success and refinement of foundation models, especially for LLMs \cite{kumar2025llm}, which refine models to improve reasoning, address limitations, and better align with user intents and ethical considerations. Among these methods, fine-tuning \cite{yue2023disc,chenlonglora,luong2024reft,ding2023parameter,malladi2023fine,xu2023parameter,guo2024tuning}, integration of RL \cite{ouyang2022training,yuan2023rrhf,liu2024skywork,dong2024rlhf, cui2023ultrafeedback,du2023guiding,song2023self,xu2025magpie}, and test-time scaling \cite{jiang2014searching,yao2023tree,yao2023tree,tian2024toward,gandhi2024stream} have been the most prominent strategies for optimizing LLMs' performance. Integrating these post-training strategies presents unique research opportunities for BFMs. For instance, leveraging RL techniques like RLHF and RLAIF for BFMs can be crucial for refining the alignment between the agent's behavior and real-world human expectations, especially in human-centric task environments. This opens avenues for developing more robust models that can adapt dynamically to user feedback. Additionally, test-time scaling for BFMs can enhance computational efficiency during deployment, particularly for real-time robot control or decision-making systems. Research could focus on improving the scalability of BFMs while ensuring that model outputs remain accurate and contextually appropriate across varying operational conditions. 

\subsubsection{Multi-agent System}
A multi-agent system (MAS) consists of multiple interacting agents that can be either cooperative, competitive, or a mix of both, aiming to address complex tasks that require collaboration, decision-making, and behavior coordination among agents \cite{arai2002advances,gautam2012review,dorri2018multi,madridano2021trajectory,zhou2021survey,ju2022review}. BFMs can fundamentally accelerate the construction of MAS consists of humanoid robots, eliminating the need to effortlessly teach each robot basic survival skills like balance and locomotion before they can collaborate. Instead, researchers can focus directly on higher-level coordination challenges, such as role allocation and team strategy. However, current BFMs trained on single-robot data lack the specialized interaction capabilities needed for optimal collaboration. This presents a promising research direction: developing next-generation BFMs trained explicitly on multi-robot interaction scenarios. Such models could better handle physical coordination challenges, such as object handovers, formation maintenance, and collision avoidance, while preserving their generalizability.

\subsubsection{Evaluation Mechanism}
While foundation models like LLMs benefit from well-established benchmarks (\textit{e.g.}, GPQA \cite{rein2024gpqa} for broad knowledge recall, MATH \cite{hendrycks2measuring} for mathematical problem solving, or MUSR \cite{spraguemusr} for multi-step reasoning), there is no specific and comprehensive evaluation mechanism for evaluating BFM's capability and guiding the evolution direction \cite{xu2024lvlm,huang2025causality}. A robust assessment of BFMs must consider multiple interdependent factors, including task generalization across unseen scenarios, adaptability to new skills with minimal data, robustness against physical perturbations, and alignment with human safety and interpretability standards. For example, Motivo \cite{tirinzoni2025zeroshot} combines quantitative metrics, such as task success rates, with qualitative human evaluations of motion naturalness; yet, critical gaps remain in assessing compositional skill combinations, hardware-specific constraints, and long-term behavioral stability. Future benchmarks should focus on progressive difficulty levels and cross-domain transfer tests to effectively assess the potential of BFMs as general-purpose physical controllers in both simulated and real-world scenarios. This approach will ultimately guide the field toward developing more capable and reliable humanoid systems.

\subsection{Risks}
\subsubsection{Ethical Issues}
Ethical issues consistently accompany the development of diverse foundation models \cite{weidinger2021ethical,yan2024practical,bommasani2024considerations}, which involve biased or unlicensed data, racial discrimination, and uncontrollable behaviors, etc. For BFMs, training on non-diverse motion datasets may encode demographic biases such as favoring movements natural to specific age groups or body types, which then propagate into robotic behaviors and create embodied forms of discrimination. Meanwhile, privacy risks escalate beyond data memorization to movement analytics, where rehabilitation or performance data could leak sensitive health information through generated motions. The physical instantiation of BFMs introduces unprecedented risks: unlike purely digital models, misaligned BFMs might reproduce unsafe or socially harmful behaviors (\textit{e.g.}, aggressive gestures, or exclusionary motions) with real-world consequences. While techniques like differential privacy and federated learning offer partial solutions, they struggle with the temporal nature of continuous motion data. In summary, BFMs demand novel governance frameworks that address both data provenance and normativity of real-time behavior.

\subsubsection{Safety Mechanism}
As BFMs will be increasingly deployed in real-world robotic systems, they introduce critical safety requirements beyond those of digital foundation models \cite{vasic2013safety,lasota2017survey,kumagai2019toward}. A key issue is maintaining model behavior integrity, particularly in safety-critical scenarios such as human-robot interaction and autonomous navigation. When trained on large-scale yet weakly curated motion datasets, BFMs may unintentionally learn unsafe or undesirable behaviors. Even minor changes in sensory input, caused by adversarial attacks or sensor noise, can lead to control failures. This highlights the need for robustness against shifts in data distribution and protection against malicious input manipulation. 

While most current BFMs focus primarily on proprioceptive inputs, integrating multimodal information (\textit{e.g.}, visual, linguistic, and auditory cues) has emerged as a promising direction for more generalizable and situationally aware control. However, multimodality introduces new vulnerabilities. Adversaries can exploit inconsistencies across modalities, as seen in the well-known CLIP case where an apple image was misclassified as an "iPod" due to an overlaid label \cite{goh2021multimodal}. Such cross-modal confusion can be especially dangerous when a BFM is adapted for an unimodal task but retains sensitivity to irrelevant signals from other modalities. These challenges underscore the need to develop robust safety mechanisms for BFMs. Future work could prioritize adversarial robustness, cross-modal consistency checks, and disentangling modality-specific information to ensure predictable, trustworthy robot behavior in open-world environments.

\section{Conclusion}\label{sec:conclusion}
In this paper, we present a systematic review of the behavior foundation model, an emerging and transformative paradigm of humanoid whole-body control systems. By pre-training on large-scale and diverse humanoid behavior data, BFMs learn broad behavior priors that enable few-shot or zero-shot adaptation to diverse downstream tasks, eliminating the need for resource-intensive and task-specific retraining. We establish a comprehensive taxonomy that categorizes BFM approaches into pre-training pipelines and adaptation strategies, demonstrating their real-world applicability across industrial, gaming, and healthcare domains. Furthermore, we highlight future research opportunities in multimodal BFMs, high-level ML system integration, post-training optimization, and standardized evaluation mechanisms that could accelerate BFM development.

Despite their unprecedented capabilities, BFMs face significant challenges, including the Sim2Real gap, embodiment dependence, and data scarcity. The physical instantiation of BFMs introduces unique safety risks that require robust verification mechanisms, while their training on human motion data likely raises ethical concerns regarding privacy and bias mitigation. Addressing these limitations in future work will lead to more reliable and generalizable BFMs. Our work is expected to inspire more subsequent research on BFMs.

\section*{Acknowledgments}
This work is supported, in part, by the Hong Kong SAR Research Grants Council under Grant No. PolyU 15224823, the Guangdong Basic and Applied Basic Research Foundation under Grant No. 2024A1515011524, the NSFC under Grant No. 62302246, the ZJNSFC under Grant No. LQ23F010008, the Ningbo under Grants No. 2023Z237 \& 2023CX050011 \& 2024Z284 \& 2024Z289 \& 2025Z038, and the Ningbo Institute of Digital Twin (IDT) under Grant No. S203.2.01.32.002. This work is also supported by the High Performance Computing Center at the IDT and the Eastern Institute of Technology, Ningbo.

\bibliography{ref}
\bibliographystyle{ieeetr}

\end{document}